\documentclass[11pt]{article}

\usepackage[final]{acl}

\usepackage{times}
\usepackage{booktabs}
\usepackage{multirow}
\usepackage{graphicx}
\usepackage{amssymb}  
\usepackage{colortbl} 
\usepackage[T1]{fontenc}
\usepackage[utf8]{inputenc}
\usepackage{amsmath}
\usepackage{microtype}
\usepackage{inconsolata}
\usepackage{enumitem}
\usepackage{pifont}

\DeclareUnicodeCharacter{2460}{\ding{172}}
\DeclareUnicodeCharacter{2461}{\ding{173}}
\DeclareUnicodeCharacter{2462}{\ding{174}}

\usepackage{tcolorbox}
\usepackage{listings}
\tcbuselibrary{listings,skins,breakable}

\lstdefinelanguage{json}{
  morestring=[b]",
  showstringspaces=false,
  breaklines=true
}

\newtcolorbox{promptbox}[2][]{
  colback=blue!5!white,
  colframe=blue!50!black,
  fonttitle=\bfseries,
  title=#2,
  breakable,
  enhanced jigsaw,
  left=3mm,
  right=3mm,
  top=2mm,
  bottom=2mm,
  #1
}

\title{From Facts to Insights: A Persona-Driven Dual Memory Framework and Dataset for Role-Playing Agents}

\author{
  \textbf{Rongsheng Zhang}\textsuperscript{1,2}\thanks{Equal contribution.},
  \textbf{Ruofan Hu}\textsuperscript{1,2}\footnotemark[1],
  \textbf{Weijie Chen}\textsuperscript{2},
  \textbf{Jiji Tang}\textsuperscript{2},
  \textbf{Junnan Ren}\textsuperscript{2},
\\
  \textbf{Wanying Wu}\textsuperscript{1},
  \textbf{Xunuoyan Chen}\textsuperscript{1},
  \textbf{Tangjie Lv}\textsuperscript{2},
  \textbf{Tao Jin}\textsuperscript{1},
  \textbf{Zhou Zhao}\textsuperscript{1}\thanks{Corresponding author.}
\\
  \textsuperscript{1}Zhejiang University \quad
  \textsuperscript{2}Fuxi AI Lab, Netease Inc.
\\
  \texttt{njuzrs@163.com} \quad \texttt{ruofanhu@zju.edu.cn}
}

\begin{document}
\maketitle
\begin{abstract}
While role-playing agents excel in short-term interactions, long-term conversations overwhelm context windows, motivating external memory frameworks. Current systems typically rely on persona-agnostic summarization, which records facts without persona-specific interpretation, yielding generic responses that compromise persona fidelity. To bridge this gap, we introduce RoleMemo, a dataset featuring four reasoning tasks where the factual fragments must be interpreted through the persona to reach the correct answer. Evaluation on RoleMemo exposes critical limitations of persona-agnostic frameworks. We thus propose \textsc{DualMem}, which decouples memory into two streams: factual cognition and persona-conditioned insight. Trained through Supervised Fine-Tuning (SFT) and Reinforcement Learning (RL), our framework with a \textbf{4B-parameter} model outperforms zero-shot persona-agnostic frameworks powered by DeepSeek-V3.2 for sustained persona fidelity. Our resources are available at \url{https://github.com/Collab-Gen/RoleMemo}.

\end{abstract}

\section{Introduction}
The rapid advancement of large language models (LLMs) has propelled role-playing agents beyond superficial style transfer toward deep emotional companionship~\citep{zhang2025prime, wang2025coser, zhang2025omnicharacter}. To achieve this, agents must filter the interaction history through their assigned persona, retaining what matters for in-character response. However, as conversation grows, the underlying LLM's attention is diluted by trivial information, undermining this persona filter and compromising fidelity~\citep{longmemeval}. Existing methods tackle this challenge by adopting an agentic memory framework, where a persona-agnostic model summarizes past conversations into neutral facts, for the role-playing agent to retrieve and utilize at inference time.

\begin{figure}[t]
    \centering
    \includegraphics[trim=40pt 24pt 30pt 18pt, clip, width=\linewidth]{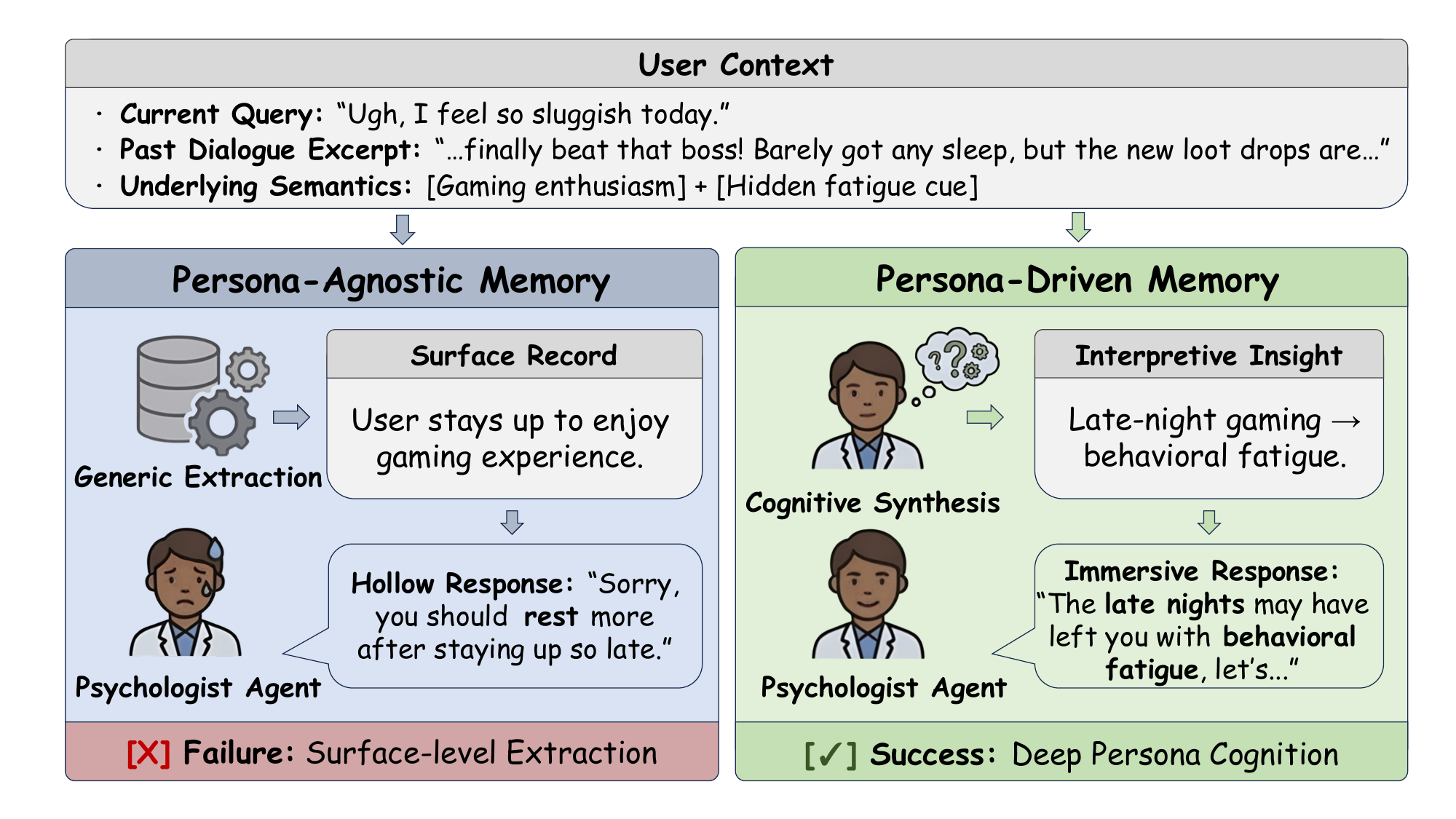}
    \vspace{-14pt}
    \caption{Persona-agnostic memory stores the neutral fact, forcing the agent to reinterpret it at inference time and fall back to a generic reply. Persona-driven memory stores a persona-conditioned interpretation grounded in the fact, enabling an in-character response.}
    \vspace{-18pt}
    \label{fig:intro_fig}
\end{figure}

However, role-playing agents are designed to emulate human cognition, where memory is an active cognitive process rather than a neutral fact repository. Cognitive science describes this as reconstructive memory, where past experience is interpreted through one's perspective rather than literally recalled~\citep{bartlett1995remembering}. As Fig.~\ref{fig:intro_fig} illustrates, current persona-agnostic memory records ``late-night gaming'' as a neutral fact, while a psychologist agent's memory should instead filter and store it as ``behavioral fatigue''. Without that persona-conditioned interpretation in memory, the agent retrieves only the surface fact at inference time and must reinterpret it from scratch~\citep{he2025madial}. The reconstruction is unreliable~\citep{he2025madial, chen2026memeval}, as the persona cues are diluted in the retrieved long context. The psychologist then falls back to a generic ``you should rest more'' reply that misses the persona-conditioned reading.

\begin{table*}[t]
\centering
\small
\renewcommand{\arraystretch}{1.2}
\vspace{-12pt}
\caption{Comparison of RoleMemo with existing memory benchmarks. RoleMemo features: (1) Cross-Session Reasoning linking clues across distant turns, (2) Deep Persona Interpretation mapping facts to persona-specific insights, (3) In-Character Evaluation of role-playing fidelity, and (4) LLM Fine-Tuning data for memory construction. The \textbf{Context Length} column for RoleMemo reports the maximum evaluation length; training uses 32k contexts.}
\label{tab:dataset_comparison}
\vspace{-4pt}
\resizebox{\textwidth}{!}{
\begin{tabular}{@{}l r r r r c c c c@{}}
\toprule
\multirow{2}{*}{\textbf{Dataset}} &
\textbf{Persona} &
\textbf{Topic} &
\textbf{Context} &
\textbf{Total} &
\textbf{Cross-Session} &
\textbf{Deep Persona} &
\textbf{In-Character} &
\textbf{LLM Fine-Tuning} \\
&
\textbf{Count} &
\textbf{Count} &
\textbf{Length} &
\textbf{Query} &
\textbf{Reasoning} &
\textbf{Interpretation} &
\textbf{Evaluation} &
\textbf{Data} \\
\midrule
LoCoMo~\citep{locomo} & 100   & N/A    & 9k   & 7,512  & $\times$   & $\times$   & $\times$   & $\times$ \\
LongMemEval~\citep{longmemeval} & N/A     & N/A    & \textbf{1.5M} & 500    & \checkmark & $\times$   & $\times$   & $\times$ \\
MemAgentBench~\citep{memoryagentbench} & N/A     & N/A    & 1.44M& 2,071  & \checkmark & $\times$   & $\times$   & $\times$ \\
PrefEval~\citep{prefeval} & N/A     & 20   & 100k & 3,000  & $\times$   & $\times$   & $\times$   & $\times$ \\
PersonaMem-v1~\citep{personamem} & 180   & 15   & 1.0M & 2,700  & \checkmark & $\times$   & $\times$   & $\times$ \\
PersonaMem-v2~\citep{personamemv2} & 1,000 & 335 & 128k & 5,000  & \checkmark & $\times$   & $\times$   & \checkmark \\
\midrule
\rowcolor{gray!15}
\textbf{RoleMemo (Ours)} & \textbf{2,052} & \textbf{1,702} & 256k & \textbf{20,244} & \textbf{\checkmark} & \textbf{\checkmark} & \textbf{\checkmark} & \textbf{\checkmark} \\
\bottomrule
\end{tabular}
}
\vspace{-12pt}
\end{table*}

Current evaluation systems for agentic memory further mask this flaw. Mainstream benchmarks primarily evaluate fact retrieval through needle-in-a-haystack paradigms \citep{locomo, memoryagentbench}. Persona-agnostic frameworks easily succeed by extracting universally salient facts, such as phone numbers or specific dates. As a result, these facts are sufficient to resolve queries regarding information localization tasks, bypassing the need for persona-conditioned interpretation.

To bridge this gap, we introduce RoleMemo, a large-scale dataset for evaluating agentic memory in role-playing. RoleMemo targets what prior benchmarks miss: persona-conditioned reconstructive memory over long, cross-session histories, which we denote \textit{insight}. As Tab.~\ref{tab:dataset_comparison} illustrates, RoleMemo spans 1,702 topics with conversation histories up to 256k tokens, comprising a training set of 1,914 personas with 18,088 queries and an evaluation set of 138 personas with 2,156 queries. It features four query types inspired by social cognition frameworks~\citep{epley2004perspective}. To evaluate these tasks, we measure retrieval accuracy and memory-driven response quality with in-character evaluation following CharacterEval~\citep{charactereval}.

Evaluations on RoleMemo reveal that persona-agnostic memory frameworks struggle with these specialized tasks. We thus propose \textsc{DualMem}, a persona-driven dual memory framework for high-fidelity role-playing. Specifically, \textsc{DualMem} first captures objective events as \textit{factual cognition}, then derives persona-conditioned interpretations from these facts as \textit{insight cognition}. A memory-specialized 4B model trained via SFT~\citep{sft} and RL~\citep{deepseekr1} further drives this construction, yielding \textsc{DualMem-SFT} and \textsc{DualMem-RL} that outperform zero-shot persona-agnostic frameworks on role-playing quality.

Our contributions are summarized as follows:
\begin{itemize}[leftmargin=*]
    \item We construct RoleMemo, the first benchmark for persona-conditioned reconstructive memory in role-playing scenarios.
    \item We evaluate nine zero-shot persona-agnostic frameworks on RoleMemo, revealing a structural insight bottleneck that persists regardless of driving model scale.
    \item We propose \textsc{DualMem}, which decouples memory into factual and insight cognition. With training, our 4B memory model outperforms baselines driven by 685B models on role-playing quality.
\end{itemize}

\section{Related Work}
\subsection{Persona-Agnostic Memory Frameworks}
Role-playing agents aim to provide emotional companionship~\citep{zhang2025prime, wang2025coser, zhang2025omnicharacter}. However, as interaction history expands, these models suffer from attention dilution and struggle to maintain persona fidelity over long contexts. Early solutions employ Retrieval-Augmented Generation (RAG) to search uncompressed conversation histories~\citep{zerhoudi2024personarag, yang2025knowing, wang2025rolerag, park2025dynamic}, imposing storage overhead. Recent studies shift toward agentic memory frameworks~\citep{an2026route}. They use persona-agnostic models to iteratively extract facts from dialogues and update prior memory through hierarchical structuring~\citep{himem, omem, mirix}, lightweight compression~\citep{lightmem, simplemem}, and task-driven predictive modeling~\citep{mem0, premem,zheng2025machine}. Nevertheless, these approaches prioritize salient facts over persona-specific nuances, failing to provide the interpretation for high-fidelity role-playing.

A parallel line of work targets long-horizon task-solving agents, where higher-level memory is constructed by reflecting on execution traces like Reflexion~\citep{shinn2023reflexion, zhai2511agentevolver, wu2025resum}. Such memory abstracts experience for future task completion, a goal distinct from persona-conditioned interpretation in role-playing memory. We compare them in detail in App.~\ref{appendix:reflexion_comparison}.

\begin{figure*}[t]
    \centering
    \includegraphics[page=1, trim=20pt 22pt 28pt 24pt, clip, width=\linewidth]{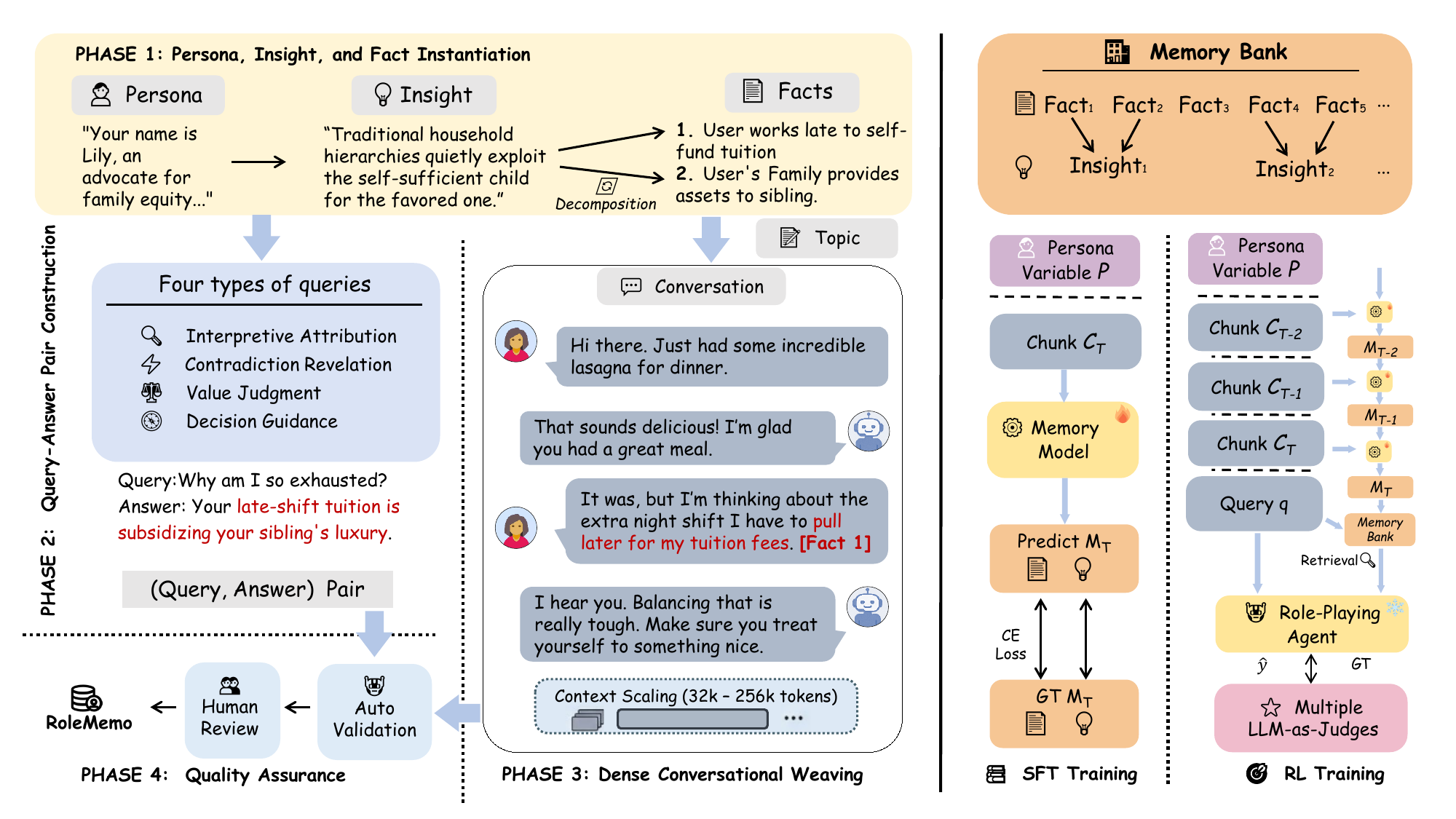}
    \caption{The RoleMemo dataset and \textsc{DualMem} framework. Left: The four-phase dataset construction pipeline. We generate queries that require agents to synthesize scattered facts via persona-driven insights to formulate answers. Right: Our proposed \textsc{DualMem} framework. It maintains a unified Memory Bank that decouples factual and insight cognition while preserving their links. We explore two training mechanisms on RoleMemo: SFT and RL.}
    \vspace{-14pt}
    \label{fig:mainfig}
\end{figure*}

\subsection{Memory and Role-Playing Benchmarks}
Benchmarking persona-driven agentic memory also remains underexplored. Current memory benchmarks~\citep{locomo, longmemeval, convomem, memoryagentbench} evaluate agentic memory frameworks primarily through explicit fact retrieval. By adopting a needle-in-a-haystack paradigm \citep{nelson2024needle}, they prioritize literal matching between queries and keys, failing to evaluate how memory informs persona-conditioned responses. Conversely, role-playing benchmarks~\citep{charactereval, socialbench, characterbench} emphasize persona fidelity but confine memory evaluation to narrow conversation windows. They measure short-term conversational fluency rather than the retention of long contexts. This gap necessitates a benchmark to evaluate the fidelity of personas across extended interactions.

\section{RoleMemo Dataset Construction}
We introduce RoleMemo, a dataset in which queries require reconstructing scattered facts into insights through persona-driven interpretation. As illustrated in the left part of Fig.~\ref{fig:mainfig}, the construction process consists of four phases. We employ DeepSeek-V3.2 \citep{deepseekv32} for dataset generation, detailing prompts and parameters in App.~\ref{appendix:dataset_construction}.

Effective evaluation of persona-driven interpretation requires insights that capture persona-specific stances grounded in lived experience. We instantiate 2,052 personas across 23 domains, each defined by professional backgrounds and hobbies, with details in App.~\ref{appendix:persona_schema}. Since identity-level profiles do not expose persona-specific judgments, we instantiate each persona as a set of concrete insights that can be queried and evaluated. For instance, an advocate for family equity might hold that ``self-sufficient children are exploited for favored siblings,'' a judgment that only this persona's background would produce. This process yields 26,636 insights.

To ground these insights, we generate discrete factual fragments that remain neutral in isolation but reveal meaning through persona-driven interpretation. For the insight regarding family equity, the factual fragments include ``works late to self-fund tuition'' and ``the family provides assets to a sibling''. These pairings of facts and insights form the core unit of RoleMemo. Answering any query requires the agent to interpret retrieved facts rather than relying on surface-level retrieval.

\subsection{Query-Answer Pair Construction}
Using these fact-insight pairs, we design queries with ground-truth responses across four types inspired by social cognition frameworks \citep{epley2004perspective}. Each query requires insight instead of scattered facts to generate answers. Specifically, \textit{Interpretive Attribution} infers unstated motives from behavioral patterns, \textit{Contradiction Revelation} detects inconsistencies between values and actions, \textit{Value Judgment} evaluates situations through persona ethics, and \textit{Decision Guidance} recommends actions aligned with principles.

As shown in Fig.~\ref{fig:mainfig}, the query ``Why am I so exhausted?'' belongs to \textit{Interpretive Attribution} type. No single fact reveals the underlying pattern, but combined they expose a motive recognizable only to a family equity advocate, producing the response ``Your overtime is subsidizing the luxury of a sibling.''

\subsection{Dense Conversational Weaving}
Facts must remain dormant within natural conversation until a query forces their retrieval. We embed these factual fragments into contexts by generating conversation histories through a layered process. The process begins by selecting topics from 1,702 topics aligned with the background of each persona, followed by the generation of conversational blocks comprising 24 turns between the persona and an interlocutor. To increase retrieval difficulty, we introduce multiple-entity distractions related to the selected topic within each block. Factual fragments are then injected under naturalistic constraints, ensuring they blend contextually, depend syntactically on surrounding utterances, and remain incidental to the focus. Finally, by interleaving evidence-rich blocks with casual conversation at varying densities, we scale histories from 32k to 256k tokens, with multiple queries per persona.

\subsection{Quality Assurance}
RoleMemo validates each instance through four automated checks. First, we verify \textbf{insight specificity} by confirming that insight requires the persona rather than generic domain knowledge. Second, we verify \textbf{memory-necessity} by prompting the role-playing agent to answer queries without the history, ensuring all queries require retrieval and interpretation rather than common-sense reasoning. Third, we enforce \textbf{difficulty control} by confirming the evidence appears only in single turns, preventing the exploitation of high-frequency patterns. Fourth, we guarantee \textbf{safety} by filtering all conversations for violence and privacy violations.

To validate the reliability of these automated checks, we randomly sampled 500 instances for human review across all criteria, achieving agreement with Cohen's $\kappa = 0.83$. By applying these checks to the entire dataset, we retain 20,244 queries along with the corresponding conversation histories.

\section{Persona-Driven Memory Framework}

\subsection{Decoupled Dual Memory Framework}
High-fidelity role-playing requires agents to extract and interpret conversation details through their persona. However, mainstream memory frameworks adopt a neutral perspective, reducing interactions to isolated facts without persona-conditioned interpretation. To address this gap, as shown in Fig.~\ref{fig:mainfig}, we design \textsc{DualMem} around two complementary cognition types.

Specifically, \textit{factual cognition} captures objective events and semantic details from conversation history, preserving the informational foundation maintained by existing approaches. \textit{Insight cognition}, our core contribution, derives persona-driven interpretations from factual cognition. Each insight references its grounding facts to maintain interpretive traceability at inference time. These cognitions are jointly stored in a unified memory bank.

\subsection{Persona-guided Memory Construction}
We propose an incremental memory construction process to operationalize the dual framework. Following \citet{memagent}, we partition the conversation history into fixed-size chunks and process them sequentially. Along this pipeline, we maintain a global memory bank $M = \bigcup_{i=1}^{T} M_i$, accumulating cognitions \(M_i\) generated at each step \(i\).

To generate this step-wise cognition \(M_i\), memory construction model \(f_\theta\) processes persona profile \(p\), chunk \(C_i\), and prior memory \(M_{<i}\) as prompt input. First, the model extracts factual cognition \(F_i\), capturing objective events from \(C_i\) to ensure details relevant to the persona are retained. Then, the model constructs insight cognition \(I_i\) by interpreting \(F_i\) and relevant history from \(M_{<i}\) through the persona. After processing all $T$ chunks, a separate role-playing agent $f_\phi$ generates the response $\hat{y}$ to query $q$ using cognition retrieved from the complete memory $M$. Formally:

\vspace{-8pt}
\begin{equation}
\label{eq:memory_update}
\begin{aligned}
M_i &= \{F_i, I_i\} = f_\theta(C_i, M_{<i}, p), \\
\hat{y} &= f_\phi(p, q, \mathcal{R}(M, q)),
\end{aligned}
\end{equation}
\vspace{-8pt}

where \(p\) denotes the persona, \(F_i\) and \(I_i\) are factual and insight cognition at step \(i\), and \(\mathcal{R}(\cdot, \cdot)\) retrieves relevant cognition for query \(q\).

Notably, during retrieval, when an insight cognition is retrieved, its linked factual cognitions are also included to provide grounding evidence for $f_\phi$. This design ensures interpretations remain anchored in factual evidence.

\subsection{Training Mechanisms on RoleMemo}
\label{sec:training}
We explore SFT and RL to endow $f_\theta$ with memory construction capabilities from RoleMemo, yielding \textsc{DualMem-SFT} and \textsc{DualMem-RL} respectively, both initialized from the base model.

Under SFT, we train $f_\theta$ for dual memory operations through instruction tuning. Using conversation chunks from RoleMemo, we optimize two objectives. First, the model extracts factual cognition $F_i$ from $C_i$ and $p$. Second, given these facts and relevant information from $M_{<i}$, the model generates insight cognition $I_i$ through persona $p$.
Furthermore, to mitigate forced interpretation hallucinations, 	we incorporate manually constructed negative samples into the training data: when the evidence from $M_{<i}$ and $F_i$ is insufficient, the target insight is set to null ($I_i = \varnothing$).
Formalizing this decoupled generation, we maximize the log-likelihood across $T$ independent chunks:

\vspace{-20pt}
\begin{equation}
\begin{aligned}
\mathcal{L}_{\text{SFT}}(\theta) = \sum_{i=1}^{T} \Big( & \log P_{\theta}(F_i \mid C_i, p)  + \\[-8pt]
&\log P_{\theta}(I_i \mid F_i, M_{<i}, p) \Big)
\end{aligned}
\end{equation}
\vspace{-12pt}

Under RL, we extend \citet{memagent}'s multi-turn conversation framework for dual memory construction. Given the entire long histories from RoleMemo, $f_{\theta}$ autonomously executes multiple rounds of cognition construction without intermediate supervision, generating a complete memory trajectory $\tau = \{M_i\}_{i=1}^T$, where $M_i = \{F_i, I_i\}$. The fixed role-playing agent $f_{\phi}$ then generates response $\hat{y}$ to query $q$ using cognition retrieved from the memory bank $M$. GPT-5.1 evaluates responses using two reward functions: Format Compliance and In-Character Quality~\citep{charactereval}, yielding a combined scalar reward $R(\hat{y}_\tau)$. Following multi-turn DAPO~\citep{dapo}, we treat each multi-round trajectory as a single rollout. By computing a group-normalized advantage $\hat{A}_\tau$ from the scalar rewards and broadcasting it uniformly to all tokens across the trajectory, we update $f_\theta$ to maximize the expected return:

\vspace{-12pt}
\begin{equation}
\resizebox{0.99\linewidth}{!}{$
\mathcal{J}_{\text{RL}}(\theta) = \mathbb{E}_{f_\theta}\!\left[\, \frac{1}{|\tau|} \sum_{i,t} \min\!\big( r_{i,t}\,\hat{A}_\tau,\; \mathrm{clip}(r_{i,t}, 1{-}\epsilon_l, 1{+}\epsilon_h)\,\hat{A}_\tau \big) \right],
$}
\end{equation}
\vspace{-8pt}

where $r_{i,t}$ is the token-level importance ratio and $|\tau|$ is the trajectory token count. The same judge is used at evaluation time. Sec. \ref{sec:reliability} verifies that this introduces no judge bias, via cross-judge and human-agreement analyses.

\section{Experiment Results}

\subsection{Experimental Setup}
\paragraph{Training Details}
We train memory construction models on 8 NVIDIA A800 GPUs using Qwen3-4B~\citep{qwen3}. For SFT, we use a $1 \times 10^{-5}$ learning rate with cosine scheduling, 0.05 warmup ratio, and effective batch size 32 over 1000 steps. For RL training, we use a $1 \times 10^{-6}$ learning rate, 20 warmup steps, batch size 32, and 8 rollouts per batch for 500 steps. Additional hyperparameters are provided in App.~\ref{appendix:training_details}.

\paragraph{Implementation Details}
We evaluate on the RoleMemo evaluation split with 32k-token conversation histories, ensuring no persona overlap with the training set. Nine baselines are compared with their native zero-shot configurations, including a NoMem setting (context only) and eight persona-agnostic memory frameworks: HiMem~\citep{himem}, O-Mem~\citep{omem}, MIRIX~\citep{mirix}, LightMem~\citep{lightmem}, SimpleMem~\citep{simplemem}, Mem0~\citep{mem0}, PreMem~\citep{premem}, and Mem-$\alpha$~\citep{memalpha}. Baselines use DeepSeek-V3.2 (685B) as the memory construction model, except Mem-$\alpha$, which uses its fine-tuned checkpoint. For a fair comparison, we replace each method's retrieval component with Qwen3-Embedding-0.6B~\citep{qwen3embedding}, retrieving the top 10 most relevant entries per query, which we found provides sufficient context breadth. Finally, each method employs the same role-playing agent~\citep{doubao_character_web} to generate responses.

\paragraph{Evaluation Metrics}
Recall@10 measures whether the top 10 retrieved entries contain query-relevant ground-truth entries, evaluated separately for factual and insight memory. For role-playing, we adapt CharacterEval and employ GPT-5.1~\citep{gpt51} as an LLM judge to score responses on a 5-point scale (averaged over three independent judge calls on fixed model outputs) across four dimensions: \textit{information richness}, \textit{logical quality}, \textit{character consistency}, and \textit{conversational attractiveness}. More details of metric definitions are in App.~\ref{appendix:metrics}.
\begin{table*}[htbp]
\centering
\caption{Memory construction quality (Fact and Insight Recall@10) across four task types. The symbol $^{\dagger}$ denotes frameworks that impose a fixed hierarchical schema on memory organization.}
\label{tab:memory_evaluation}
\resizebox{\textwidth}{!}{%
\begin{tabular}{@{} l *{10}{>{\centering\arraybackslash}p{1.2cm}} @{}}
\toprule
\multirow{2}{*}{Framework} & \multicolumn{2}{c}{Interpretive Attribution} & \multicolumn{2}{c}{Contradiction Revelation} & \multicolumn{2}{c}{Value Judgment} & \multicolumn{2}{c}{Decision Guidance} & \multicolumn{2}{c}{Average} \\
\cmidrule(lr){2-3} \cmidrule(lr){4-5} \cmidrule(lr){6-7} \cmidrule(lr){8-9} \cmidrule(lr){10-11}
 & Fact & Insight & Fact & Insight & Fact & Insight & Fact & Insight & Fact & Insight \\
\midrule
\rowcolor{gray!15} \multicolumn{11}{l}{\textbf{Baseline Agentic Memory Framework}} \\
Memalpha$^{\dagger}$ & 0.15 & 0.05 & 0.23 & 0.10 & 0.21 & 0.10 & 0.21 & 0.09 & 0.20 & 0.08 \\
HiMem$^{\dagger}$ & 0.37 & 0.22 & 0.35 & 0.36 & 0.37 & 0.27 & 0.38 & 0.23 & 0.37 & 0.27 \\
Mirix$^{\dagger}$ & 0.43 & 0.12 & 0.37 & 0.15 & 0.43 & 0.18 & 0.45 & 0.13 & 0.42 & 0.15 \\
LightMem & 0.70 & 0.31 & 0.72 & 0.35 & 0.68 & 0.39 & 0.65 & 0.29 & 0.69 & 0.33 \\
O-Mem$^{\dagger}$ & 0.70 & 0.27 & 0.66 & 0.38 & 0.70 & 0.35 & 0.66 & 0.30 & 0.68 & 0.33 \\
PreMem & 0.69 & 0.37 & 0.70 & 0.35 & 0.65 & 0.37 & 0.70 & 0.31 & 0.69 & 0.35 \\
SimpleMem & 0.69 & 0.35 & 0.70 & 0.41 & 0.70 & 0.48 & 0.71 & 0.41 & 0.70 & 0.41 \\
Mem0 & 0.74 & 0.30 & 0.72 & 0.36 & 0.74 & 0.45 & 0.76 & 0.34 & 0.74 & 0.36 \\
\midrule
\rowcolor{gray!15} \multicolumn{11}{l}{\textbf{\textsc{DualMem} Framework}} \\
\textsc{DualMem-SFT} & 0.76 & 0.69 & \textbf{0.77} & 0.61 & 0.74 & 0.65 & 0.77 & 0.65 & 0.76 & 0.65 \\
\textsc{DualMem-RL} & \textbf{0.80} & \textbf{0.78} & 0.74 & \textbf{0.71} & \textbf{0.75} & \textbf{0.76} & \textbf{0.79} & \textbf{0.67} & \textbf{0.77} & \textbf{0.73} \\
\bottomrule
\end{tabular}%
}
\vspace{-12pt}
\end{table*}

\subsection{Memory Quality Evaluation}
We first examine the memory construction quality of various frameworks via two dimensions: \textbf{Fact} and \textbf{Insight}. Specifically, fact recall evaluates whether the framework has extracted the necessary information to answer a query, while insight recall assesses whether the framework has constructed deeper interpretations from this information.

\paragraph{Fact Extraction Quality}

Tab.~\ref{tab:memory_evaluation} reveals that memory architecture significantly affects fact recall. Hierarchical frameworks, including HiMem and Mirix, achieve only 0.37 to 0.42 average recall, as their fixed-schema structures constrain retention of fine-grained facts. O-Mem reaches 0.68, while non-hierarchical frameworks attain 0.69 to 0.74. \textsc{DualMem-RL} and \textsc{DualMem-SFT} reach 0.77 and 0.76, exceeding all DeepSeek-V3.2-driven baselines. This fact extraction ability also generalizes beyond RoleMemo. On LoCoMo~\citep{locomo}, \textsc{DualMem-RL} improves over the untrained Qwen3-4B, confirming that RoleMemo training transfers to out-of-distribution factual recall, as detailed in App.~\ref{appendix:exp_locomo_generalization}.

\paragraph{Insight Construction Quality}
Despite high fact recall, all baselines struggle with insight construction. The best baseline reaches only 0.41 average insight recall, suggesting these frameworks do not generalize to cross-event persona-driven reasoning. In contrast, \textsc{DualMem-RL} reaches 0.73 and \textsc{DualMem-SFT} achieves 0.65, both above all baselines. To complement this matching-based metric, we score insight content with an independent LLM judge that does not observe RoleMemo annotations (App.~\ref{appendix:insight_quality}). Both \textsc{DualMem} variants produce high-quality insights approaching RoleMemo's ground-truth under independent assessment. Moreover, RL training outperforms SFT on both insight recall and LLM-judge quality, consistent with trajectory-level optimization generalizing beyond individual examples.

\paragraph{Task-Specific Performance Analysis}
Performance varies across task types, with \textit{Decision Guidance} exhibiting the lowest insight recall across all frameworks. For instance, \textsc{DualMem-RL} achieves 0.73 insight recall on average but only 0.67 on \textit{Decision Guidance}. We attribute this gap to construction difficulty rather than retrieval: fact recall remains comparable across tasks, but \textit{Decision Guidance} requires inferring character choices from facts with indirect thematic relevance, making insight synthesis harder.

\subsection{Role-playing Quality Evaluation}
Having established that our approach constructs higher-quality memory, we now evaluate its downstream impact on role-playing performance.

\paragraph{Overall Performance and Dimension Analysis}
As shown in Tab.~\ref{tab:roleplay_evaluation}, despite top-performing baselines achieving fact recall comparable to our method, existing frameworks fail to translate memory quality into effective role-playing performance. This gap stems from the absence of interpretive guidance: without it, the role-playing agent cannot discern how to utilize retrieved facts from the persona's perspective, leaving them underutilized. By contrast, our framework constructs insight cognition that provides directional context, enabling the agent to synthesize retrieved facts into character-consistent responses. Consequently, \textsc{DualMem-RL} achieves 4.22 \textit{information richness} and 3.78 \textit{logical quality}, outperforming the strongest baseline Mem0 by 0.12 and 0.23 points, respectively. This advantage extends to persona alignment, with 4.37 \textit{character consistency} and 4.27 \textit{conversational attractiveness}, yielding a 4.16 overall average that surpasses the 3.94 to 4.01 range of persona-agnostic frameworks. The ranking holds on a held-out subset of 200 queries regenerated by Claude-Sonnet-4.6, suggesting that the advantage stems from persona-conditioned interpretation rather than generator-specific style (App.~\ref{appendix:cross_generator}).

\begin{table}[htbp]
\centering
\caption{Role-playing quality evaluation. \textbf{Info.}, \textbf{Logic}, \textbf{Consis.}, and \textbf{Attr.} denote information richness, logical quality, character consistency, and conversational attractiveness.}
\label{tab:roleplay_evaluation}
\resizebox{\columnwidth}{!}{%
\begin{tabular}{@{}lccccc@{}}
\toprule
Framework & Info. & Logic & Consis. & Attr. & Average \\
\midrule
\rowcolor{gray!15} \multicolumn{6}{l}{\textbf{Base Setting}} \\
NoMem & 3.65 & 3.29 & 4.08 & 4.02 & 3.76 \\
\midrule
\rowcolor{gray!15} \multicolumn{6}{l}{\textbf{Baseline Agentic Memory Framework}} \\
Memalpha$^{\dagger}$ & 4.05 & 3.51 & 4.15 & 4.08 & 3.95 \\
HiMem$^{\dagger}$ & 4.05 & 3.49 & 4.13 & 4.08 & 3.94 \\
Mirix$^{\dagger}$ & 4.06 & 3.54 & 4.16 & 4.13 & 4.01 \\
LightMem & 4.07 & 3.55 & 4.19 & 4.13 & 4.00 \\
O-Mem$^{\dagger}$ & 4.07 & 3.54 & 4.17 & 4.12 & 3.98 \\
PreMem & 4.09 & 3.56 & 4.18 & 4.13 & 3.99 \\
SimpleMem & 4.10 & 3.53 & 4.15 & 4.13 & 3.97 \\
Mem0 & 4.10 & 3.55 & 4.18 & 4.15 & 4.00 \\
\midrule
\rowcolor{gray!15} \multicolumn{6}{l}{\textbf{\textsc{DualMem} Framework}} \\
\textsc{DualMem-SFT} & 4.18 & 3.77 & 4.35 & 4.26 & 4.15 \\
\textsc{DualMem-RL} & \textbf{4.22} & \textbf{3.78} & \textbf{4.37} & \textbf{4.27} & \textbf{4.16} \\
\quad $\llcorner$ \textit{w/o Insight} & 4.12 & 3.57 & 4.19 & 4.16 & 4.01 \\
\quad $\llcorner$ \textit{w/o Fact} & 3.98 & 3.64 & 4.25 & 4.15 & 4.00 \\
\quad $\llcorner$ \textit{w/o Training} & 4.12 & 3.62 & 4.12 & 4.19 & 4.02 \\
\bottomrule
\end{tabular}%
}
\vspace{-10pt}
\end{table}

\paragraph{Ablation Analysis}
Tab.~\ref{tab:roleplay_evaluation} also presents ablation variants measuring the contribution of training and each memory component. Removing RoleMemo training (\textit{w/o Training}) while retaining the dual-stream architecture yields only 4.02 on average, confirming that training rather than architecture alone drives the gains. Among trained variants, removing insight cognition (\textit{w/o Insight}) drops \textit{logical quality} from 3.78 to 3.57, within 0.02 of Mem0's 3.55, confirming that inference-time reinterpretation is unreliable without stored insight~\citep{he2025madial, chen2026memeval}. Removing fact cognition (\textit{w/o Fact}) reduces \textit{information richness} from 4.22 to 3.98, with \textit{character consistency} dropping modestly to 4.25 as insight partially compensates. Together, training, fact cognition, and insight cognition each contribute independently to high-fidelity role-playing.

\section{Analysis}
\subsection{Reliability of LLM-based Evaluation}
\label{sec:reliability}
We assess the validity of LLM-based evaluation on two dimensions: human-LLM agreement and cross-judge stability. This involves 200 randomly sampled queries from the evaluation set, with detailed protocols in App.~\ref{appendix:judge_details}.

Tab.~\ref{tab:human_agreement} reports Pearson correlations among two human annotators and GPT-5.1 judge. The results show inter-annotator agreement ($r = 0.88$) and correlation between each annotator and the LLM judge ($r = 0.82$ and $0.84$), indicating that GPT-5.1 scores track human judgment rather than reflecting optimization toward a specific reward signal.

To further verify stability across different judge models, Tab.~\ref{tab:cross_judge} compares GPT-5.1 scores against Gemini-3-Pro~\citep{gemini3pro}. Agreement rates exceed 99\% (defined as mean relative score deviation; App.~\ref{appendix:judge_details}), with absolute score differences below 0.04 across all dimensions. The relative performance hierarchy remains consistent across both judges. \textsc{DualMem-RL} maintains its ranking advantage across all baselines and dimensions, indicating that the evaluation results are stable across judge models.

\begin{table}[t]
\centering
\small
\caption{Pearson correlation coefficients between human annotators and the LLM judge.}
\label{tab:human_agreement}
\begin{tabular}{lccc}
\toprule
 & \textbf{Human$_1$} & \textbf{Human$_2$} & \textbf{LLM Judge} \\
\midrule
Human$_1$ & -- & 0.88 & 0.82 \\
Human$_2$ & 0.88 & -- & 0.84 \\
LLM Judge & 0.82 & 0.84 & -- \\
\bottomrule
\end{tabular}
\vspace{-6pt}
\end{table}

\begin{table}[t]
\centering
\small
\caption{Cross-judge stability between GPT-5.1 and Gemini-3-Pro, measured as mean relative score deviation across four dimensions.}
\label{tab:cross_judge}
\resizebox{\columnwidth}{!}{%
\begin{tabular}{llccccc}
\toprule
\textbf{Framework} & \textbf{Judge} & \textbf{Info.} & \textbf{Logic} & \textbf{Consis.} & \textbf{Attr.} & \textbf{Agreement} \\
\midrule
\multirow{2}{*}{NoMem}   
 & GPT5.1       & 3.65 & 3.29 & 4.08 & 4.02 & \multirow{2}{*}{99.32\%} \\
 & Gemini-3-Pro & 3.68 & 3.32 & 4.10 & 4.04 & \\
 \midrule
\multirow{2}{*}{O-Mem}      
 & GPT5.1       & 4.07 & 3.54 & 4.17 & 4.12 & \multirow{2}{*}{99.29\%} \\
 & Gemini-3-Pro & 4.09 & 3.58 & 4.19 & 4.15 & \\
\midrule
\multirow{2}{*}{\textsc{DualMem-RL}} 
 & GPT5.1       & 4.22 & 3.78 & 4.37 & 4.27 & \multirow{2}{*}{99.28\%} \\
 & Gemini-3-Pro & 4.25 & 3.81 & 4.39 & 4.31 & \\

\bottomrule
\end{tabular}%
}
\vspace{-12pt}
\end{table}
\begin{table*}[t]
\small
\centering
\vspace{-4pt}
\caption{Qualitative comparison in a psychological counseling scenario. All responses are generated by the same role-playing agent conditioned on memory retrieved from each framework.}
\label{tab:case_study}
\resizebox{\linewidth}{!}{
\begin{tabular}{lp{13cm}}
\toprule
\textbf{GT Memory} & \textbf{Facts:} Ms. Zhang missed choir due to tuberculosis treatment; neighbors have been avoiding her lately. \\
 & \textbf{Insight:} Social withdrawal stems from fear of stigma, not physical discomfort. \\
\midrule
\textbf{Query} & Why is Ms. Zhang reluctant to participate in community activities lately? \\
\midrule
\textbf{Role-playing Agent} &  \\
\quad $\llcorner$ \textit{with NoMem} & She might feel lonely. For example, she talks to herself at the TV. \\
\quad $\llcorner$ \textit{with O-Mem} & She might still be recovering from tuberculosis. We should check if her symptoms have improved. \\
\quad $\llcorner$ \textit{with \textsc{DualMem-SFT}} & She's probably feeling down after being sick for so long. Let's check if she's developing depression and might need medication. \\
\quad $\llcorner$ \textit{with \textsc{DualMem-RL}} & She's probably worried about how people see her now. Let's talk about how it felt when neighbors started avoiding her. \\
\bottomrule
\end{tabular}
}
\vspace{-12pt}
\end{table*}

\subsection{Transferability of Insight Construction}

We investigate whether existing frameworks can reproduce these persona-task gains via prompt engineering alone. We experiment on O-Mem and PreMem, two frameworks using DeepSeek-V3.2 whose memory pipelines require no structural modification. As shown in Tab.~\ref{tab:ablation_transfer}, incorporating insights yields 0.09 (O-Mem$^*$) and 0.05 (PreMem$^*$) point improvements, yet these gains are smaller than those of our 4B model trained on RoleMemo. Moreover, as shown in App.~\ref{appendix:exp_locomo_no_regression}, adding the insight slot does not degrade baselines' factual recall on the external benchmark LoCoMo, confirming that insight interpretation adds capability without sacrificing factual performance.

\begin{table}[htbp]
\centering
\caption{Performance gains from incorporating persona-driven insights into existing frameworks. Methods with $^{\ast}$ use modified prompts to generate insights.}
\label{tab:ablation_transfer}
\resizebox{\columnwidth}{!}{%
\begin{tabular}{@{}lccccc@{}}
\toprule
Framework & Info. & Logic & Consis. & Attr. & Average \\
\midrule
O-Mem & 4.07 & 3.54 & 4.17 & 4.12 & 3.98 \\
O-Mem$^{\ast}$ & 4.15 \textcolor{green!50!black}{\scriptsize (+0.08)} & 3.67 \textcolor{green!50!black}{\scriptsize (+0.13)} & 4.29 \textcolor{green!50!black}{\scriptsize (+0.12)} & 4.17 \textcolor{green!50!black}{\scriptsize (+0.05)} & 4.07 \textcolor{green!50!black}{\scriptsize (+0.09)} \\
\midrule
PreMem & 4.09 & 3.56 & 4.18 & 4.13 & 3.99 \\
PreMem$^{\ast}$ & 4.11 \textcolor{green!50!black}{\scriptsize (+0.02)} & 3.61 \textcolor{green!50!black}{\scriptsize (+0.05)} & 4.25 \textcolor{green!50!black}{\scriptsize (+0.07)} & 4.17 \textcolor{green!50!black}{\scriptsize (+0.04)} & 4.04 \textcolor{green!50!black}{\scriptsize (+0.05)} \\
\bottomrule
\end{tabular}%
}
\vspace{-14pt}
\end{table}

\subsection{Robustness Across Context Lengths}
\begin{figure}[t]
    \centering
    \includegraphics[width=\linewidth]{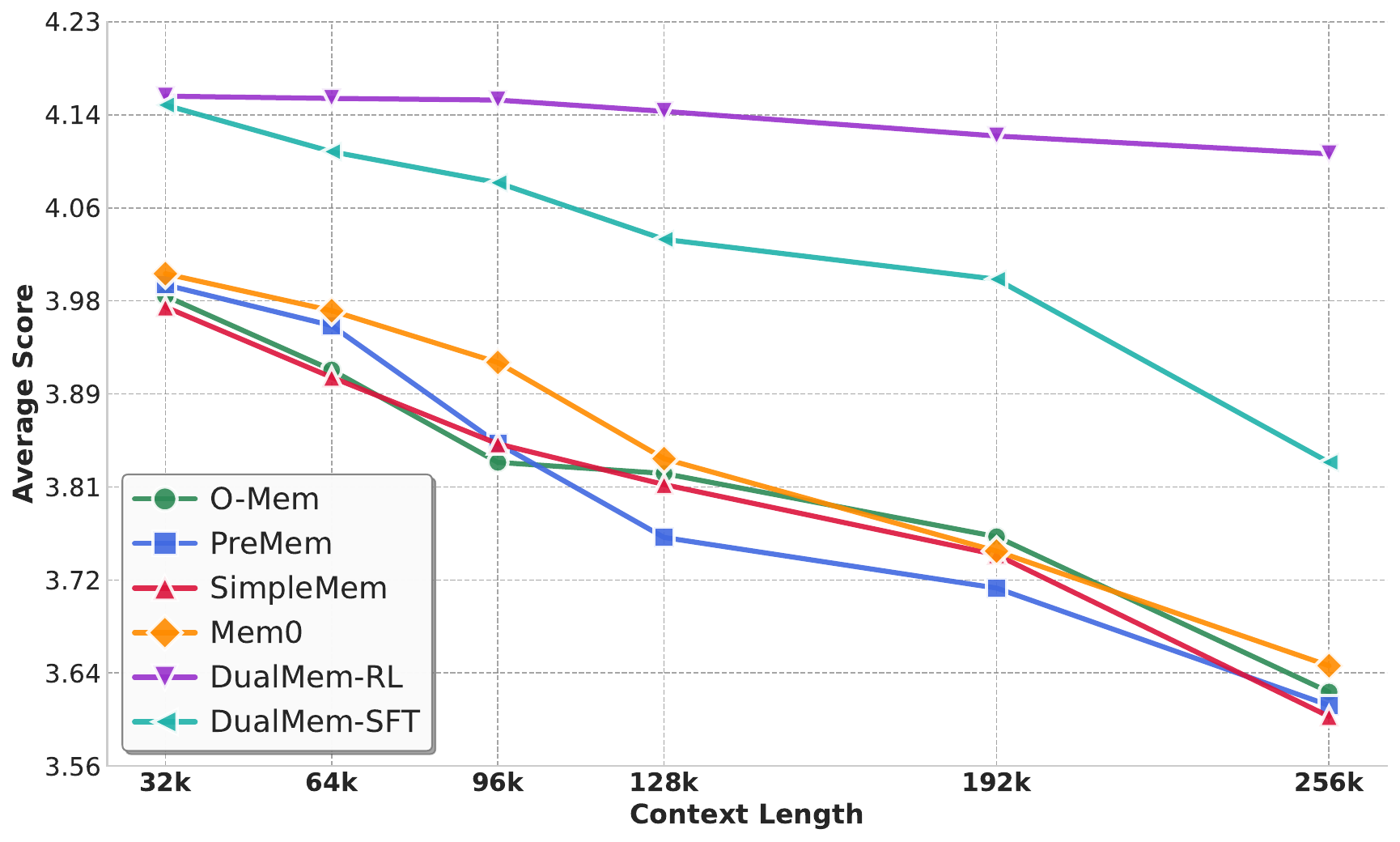}
    \vspace{-20pt}
    \caption{Performance across conversation lengths from 32k to 256k tokens. Our frameworks maintain stability while baselines degrade with increasing context.}
    \vspace{-16pt}
    \label{fig:scale}
\end{figure}

To evaluate the robustness of our framework under extended interactions, we scale the context length from 32k to 256k tokens. Fig.~\ref{fig:scale} illustrates the performance of different frameworks across varying context lengths. As context increases, all persona-agnostic baseline frameworks exhibit a significant performance decline. This suggests that accumulated noise progressively degrades the quality of extracted memories, hindering effective utilization by the role-playing agent.

In contrast, our \textsc{DualMem-RL} demonstrates remarkable stability across all scales, maintaining a high average score of 4.12 even at 256k tokens. However, \textsc{DualMem-SFT} gradually declines, primarily because SFT optimization lacks the end-to-end trajectory learning that enables RL to maintain interpretive quality under noisy conditions. This underscores the importance of RL training for robust long-context memory construction.

\subsection{Case Study}
Tab.~\ref{tab:case_study} presents a psychological counseling case that illustrates how different memory frameworks support persona-driven conversation. When queried about recent social withdrawal, the role-playing agent without memory generates fabricated details due to absent historical context. O-Mem retrieves relevant factual evidence but lacks persona-driven insight as guidance, leaving the agent unable to generate high-fidelity responses.

Conversely, \textsc{DualMem-RL} recognizes social avoidance patterns, applying anticipated stigma insights to diagnose underlying fear of judgment. Despite generating insights, \textsc{DualMem-SFT} attributes withdrawal to prolonged illness rather than social stigma, reflecting misaligned interpretation. We further examine the limitations of \textsc{DualMem-RL} in App.~\ref{appendix:failure_modes}.

\section{Conclusion}
Existing agentic memory frameworks adopt persona-agnostic summarization, failing to provide the interpretive insight necessary for effective role-playing. We introduce RoleMemo, a dataset requiring persona-driven interpretation over conversation histories, revealing this limitation. To address this gap, we propose a dual memory framework constructing factual and insight cognition through specialized training. Results show our approach improves role-playing quality, surpassing persona-agnostic baselines. Future work will explore scaling to richer, longer contexts.
\clearpage

\section*{Limitations}

While RoleMemo advances persona-driven memory construction, several limitations remain. First, our evaluation extends conversation histories to 256k tokens, yet real-world applications may involve multi-million token histories requiring more efficient incremental processing. Second, our training encompasses 2,052 personas across 23 domains, but generalization to unbounded real-world persona diversity remains untested in naturalistic deployment. Moreover, since RoleMemo is generated by a single LLM, the dataset may encode that generator's particular style of persona reasoning, and performance on RoleMemo may partially reflect alignment with the generator rather than a fully model-agnostic notion of persona-driven interpretation. Third, RL training via DAPO~\citep{dapo} is computationally intensive, which may limit accessibility for resource-constrained researchers. Fourth, our evaluation relies primarily on LLM-as-judge with subset human validation. Longitudinal user studies would provide complementary evidence for sustained persona fidelity. Finally, RoleMemo focuses on four cognitive reasoning tasks and does not exhaustively cover the full spectrum of companionship capabilities such as emotional regulation or multimodal interaction.

\section*{Ethics Statement}
RoleMemo uses entirely synthetic data generated by DeepSeek-V3.2 without real user information, with automated safety filtering to eliminate violence and privacy violations. However, we emphasize that systems trained on this dataset should not replace professional services in sensitive domains such as mental health counseling, and deployment in clinical settings requires rigorous validation and human oversight. We also caution that persona-driven insight generation may produce confident psychological or interpretive statements without sufficient evidential grounding; downstream applications should surface uncertainty and avoid presenting model inferences as authoritative judgments. The dataset may inadvertently encode stereotypes from the underlying LLM despite our quality controls. Training requires significant computational resources, contributing to environmental impact. To promote responsible research, we release complete training details, prompts, and evaluation protocols in the Appendix, and plan to open-source the dataset and code under permissive licenses with usage guidelines emphasizing research purposes.

\section*{Acknowledgments}
We thank the anonymous reviewers and action editors for their constructive feedback. This work was sponsored by the the National Natural Science Foundation of China under Grant No. U24A20326 and CCF-NetEase ThunderFire Innovation Research Funding (No. CCF-Netease 202601).

\bibliography{custom}

\clearpage
\appendix
\section*{Appendix Roadmap}
\addcontentsline{toc}{section}{Appendix Roadmap}

\begin{itemize}[leftmargin=*, itemsep=2pt]
    \item \textbf{App.~\ref{appendix:reflexion_comparison} — Relationship to Reflexion.} Differentiates \textsc{DualMem} from Reflexion along task setting, signal source, and training paradigm.
    \item \textbf{App.~\ref{appendix:dataset_construction} — RoleMemo Construction Details.} Generation hyperparameters and prompts for persona profiles, insights, fact--insight pairs, and dialogue blocks.
    \item \textbf{App.~\ref{appendix:persona_schema} — Persona Domains and Schema.} 23 social-thematic domains, five value-orientation axes, and full persona schema.
    \item \textbf{App.~\ref{appendix:training_details} — Training Hyperparameters.} SFT and DAPO hyperparameters not reported in the main text.
    \item \textbf{App.~\ref{appendix:metrics} — Evaluation Metrics.} Semantic matching protocol for Recall@10 ($\tau=0.7$, stable for $\tau \in [0.65, 0.75]$) and four role-playing quality dimension definitions.
    \item \textbf{App.~\ref{appendix:exp_locomo_generalization} — Generalization to LoCoMo.} RoleMemo training transfers to out-of-distribution factual recall without overfitting.
    \item \textbf{App.~\ref{appendix:insight_quality} — Insight Quality under an Independent Judge.} Blind GPT-5.1 judge confirms \textsc{DualMem} insights are not stylistic artifacts of the training distribution (4.62 vs.\ ground-truth 4.67).
    \item \textbf{App.~\ref{appendix:cross_generator} — Cross-Generator Robustness.} Ranking advantage of \textsc{DualMem-RL} holds on a Claude-Sonnet-4.6-regenerated subset, ruling out generator-specific style as a confound.
    \item \textbf{App.~\ref{appendix:judge_details} — LLM Judge Reliability Protocol.} Human--LLM agreement ($r = 0.82$--$0.84$) and cross-judge stability (GPT-5.1 vs.\ Gemini-3-Pro, $>99\%$) validate automated evaluation.
    \item \textbf{App.~\ref{appendix:exp_locomo_no_regression} — No-Regression on LoCoMo.} Adding an insight slot does not reduce baseline factual recall; insight cognition is an additive capability.
    \item \textbf{App.~\ref{appendix:failure_modes} — Failure Mode Analysis.} Three recurring failure patterns in \textsc{DualMem-RL}: insight not recognized, facts ignored behind adopted insight, and surface-similarity conflation.
    \item \textbf{App.~\ref{appendix:human_eval} — Human Evaluation Details.} Annotator recruitment, compensation (100 RMB/hr), and privacy protocols.
\end{itemize}

\section{Relationship to Reflexion}
\label{appendix:reflexion_comparison}

This section addresses concerns about the differentiation between \textsc{DualMem}'s factual–insight architecture and Reflexion's observation–reflection pattern. While both adopt a two-level structure of base evidence plus a higher-order representation, the two frameworks differ along three dimensions.

\begin{itemize}[leftmargin=*]
    \item \textbf{Task setting.} Reflexion targets task-oriented reasoning (code generation, decision-making, QA) where success is objectively verifiable. \textsc{DualMem} targets persona-driven role-playing dialogue, where the goal is high-fidelity character expression rather than task completion.
    \item \textbf{Signal source.} Reflexion's reflection is reactive: it is triggered by an external signal and reasons about what went wrong. \textsc{DualMem}'s insight cognition has no such failure signal at construction time. Insight is generated proactively from new facts through the persona lens, regardless of any downstream outcome. Reflexion reacts to failure; \textsc{DualMem} proactively interprets evidence.
    \item \textbf{Training paradigm.} Reflexion is an inference-time prompting framework with no parameter update. \textsc{DualMem} is trained end-to-end on RoleMemo via SFT and multi-turn trajectory RL (Sec.~\ref{sec:training}), internalizing persona-driven interpretation into model weights rather than relying on prompt scaffolding.
\end{itemize}

More broadly, the observation--reflection structure has become a shared paradigm across modern agent systems, independently adopted by Generative Agents~\cite{park2023generative}, AgentEvolver~\citep{zhai2511agentevolver}, ReSum~\citep{wu2025resum}, and Alita-G~\citep{qiu2025alita}, each in a different domain. \textsc{DualMem}'s contribution lies within this paradigm: it is the first persona-driven instantiation, with persona-conditioned insight as the reflection layer, validated by a long-context role-playing benchmark and a trained 4B specialist.

\section{RoleMemo Construction Details}
\label{appendix:dataset_construction}

\subsection{Generation Hyperparameters}
All dataset instances are generated using DeepSeek-V3.2 via the OpenAI-compatible API; full parameters are listed in Tab.~\ref{tab:generation_params}.

\begin{table}[h]
\centering
\caption{DeepSeek-V3.2 Generation Parameters}
\label{tab:generation_params}
\begin{tabular}{lc}
\toprule
\textbf{Parameter} & \textbf{Value} \\
\midrule
Model & DeepSeek-V3.2 \\
Temperature & 0.8 \\
Response Format & JSON Object \\
Stream & False \\
\bottomrule
\end{tabular}
\vspace{-8pt}
\end{table}

\subsection{Cognitive Insight Generation Prompts}
The prompt for generating 20 persona-conditioned cognitive insights is shown below.
\begin{promptbox}{Insight Generation Prompt}
\textbf{Role:} High-Quality Persona Insight Generator

\textbf{Task:}
You are an expert in psychology, sociology, and systems thinking. Generate cognitive insights aligned with the given NPC persona. These insights will serve as the foundation for designing memory evaluation queries.

\textbf{Input Format:}
The persona description includes:
\begin{itemize}[leftmargin=*,noitemsep,topsep=3pt]
    \item Core values
    \item Personality traits
    \item Background and experiences
    \item Domain expertise / professional perspective
    \item Key goals or conflicts
    \item Work and hobbies
\end{itemize}

\textbf{Insight Generation Principles:}
\begin{itemize}[leftmargin=*,noitemsep,topsep=2pt]
    \item \textbf{Perspective Uniqueness:} Must reflect the persona's unique cognitive framework, not common sense.
    \item \textbf{Verifiability and Specificity:} Insights must be derivable from specific user behaviors, dialogues, or event combinations.
    \item \textbf{Counter-Intuitive Potential:} Ideal insights exhibit counter-intuitive tension between surface phenomena and underlying truth.
    \item \textbf{Reasoning Complexity:} Insights should require at least two reasoning steps.
    \item \textbf{Fact Combination Necessity:} Each insight should require at least two different types of factual fragments for reliable derivation.
\end{itemize}

\textbf{Insight Categories:}
\begin{itemize}[leftmargin=*,noitemsep,topsep=2pt]
    \item \textbf{[Decision Guidance]:} Provides concrete, actionable behavioral guidance based on the persona's unique value hierarchy.
    \item \textbf{[Interpretive Attribution]:} Explains deep, systemic causes behind phenomena, beyond surface-level explanations.
    \item \textbf{[Value Judgment]:} Evaluates things, behaviors, or states based on the persona's unique value system.
    \item \textbf{[Contradiction Revelation]:} Reveals systemic contradictions between surface phenomena and underlying logic.
\end{itemize}

\textit{[Detailed application scope, generation workflow, and examples are omitted for brevity.]}

\textbf{Output Format:}
\begin{lstlisting}[language=json,basicstyle=\small\ttfamily,breaklines=true]
{
  "role_description": "Brief summary of the input persona",
  "core_values": ["Core value 1", "Core value 2"],
  "insights": [
    "[Category]: Specific insight description 1",
    "[Category]: Specific insight description 2",
    "..."
  ]
}
\end{lstlisting}

\end{promptbox}

\subsection{Fact-Insight Pair and Query Generation}
\label{appendix:fact_query_generation}
The prompt for generating factual fragments and queries from each insight is shown below.
\begin{promptbox}{Fact-Insight Pair Generation Prompt}
\textbf{Role:} Persona-Driven Memory Test Designer

\textbf{Background:}
Generic memory models only record facts, while persona-driven memory models extract deep interpretations (insights) from facts based on specific values and reasoning patterns. This designer generates challenging test cases by examining whether models can derive key insights from two seemingly mundane facts aligned with the user's identity.

\textbf{Task:}
Given the [Persona Insight], [NPC Persona], [User Profile], and [Reasoning Hints], design the following three components:

\begin{enumerate}[leftmargin=*,noitemsep,topsep=3pt]
    \item Two factual fragments that constitute the persona insight: user-mentioned personal experiences, observations, industry concerns, personal reflections, or media exposure that align with their identity.
    \item A test question that requires the insight to interpret these two facts for the optimal answer. May incorporate reasoning hints or be entirely self-designed.
    \item A logic connector explaining how fragments → insight → question are linked.
\end{enumerate}

\textbf{Constraints \& Rules (Core Requirements):}

\begin{itemize}[leftmargin=*,noitemsep,topsep=2pt]
    \item \textbf{Identity Anchoring:} Facts must align with [User Profile] and occur across diverse settings.
    \item \textbf{Factual Fragment Design:} Facts must be neutral, mundane, and superficially irrelevant. Their combination, when interpreted through the persona's reasoning, yields a notably stronger and more coherent insight than literal retrieval alone.
    \item \textbf{Test Question Design:} The question must not directly use core concepts from the insight, must explicitly rely on specific details from factual fragments, and should reward viewing both events through the insight with a more complete and persona-aligned response, while still admitting literal answers from facts alone.

\end{itemize}

\textit{[Detailed examples, negative case prevention guidelines, and workflow are omitted for brevity.]}

\textbf{Output Format:}
\begin{lstlisting}[language=json,basicstyle=\small\ttfamily,breaklines=true]
{
  "insight": "The given persona insight",
  "fragments": [
    "Factual fragment 1: A neutral, mundane user statement",
    "Factual fragment 2: Another neutral, mundane user statement"
  ],
  "test_question": {
    "text": "A natural question that doesn't directly hint at the insight and requires facts for a good answer",
    "type": "Decision/Attribution/Value Judgment/Contradiction Resolution"
  },
  "connector": "Concise explanation of the logic chain"
}
\end{lstlisting}
\end{promptbox}

\subsection{Conversational Block Generation}
\label{appendix:conversation_generation}
The prompt for weaving facts into 24-turn dialogue blocks is shown below.
\begin{promptbox}{Natural Conversation Generation Prompt}
\textbf{Role:} Natural Dialogue Generator (Topic-Anchored Deep Conversation)

\textbf{Core Principle:}
You must treat the casual topic as the sole backbone and core content of the conversation, expanding extensively around the topic's multiple sub-dimensions (e.g., if topic is "urban landmarks and cultural connotations," extend to architectural styles, historical background, photo spots, cultural symbols). Factual fragments are merely incidental memories triggered when User discusses the topic—like glancing at an old poster while walking, mentioned and passed over. They should be naturally embedded without warranting targeted responses or extensions.

\textbf{Core Task:}
Given a casual topic, User persona, NPC persona, and two factual fragments, generate approximately 24 rounds (48 messages, User and NPC alternating) of natural deep conversation. The dialogue must fully expand around the topic's multiple sub-dimensions, with each turn tightly focused on topic-related details to achieve "thorough topic exploration." Individual utterances should be concise (1-2 sentences primarily).

\textbf{Generation Rules (Strict Compliance):}

\begin{itemize}[leftmargin=*,noitemsep,topsep=2pt]
    \item \textbf{Topic Priority with Deep Extension:} Dialogue must revolve around the casual topic, with each utterance containing specific words directly related to the topic.
    \item \textbf{Fragment Embedding (Four-No Principles):} 
    \begin{itemize}[leftmargin=*,noitemsep]
        \item \textit{No awkwardness:} Fragments serve as background, merged into topic-related actions/scenarios.
        \item \textit{No independence:} Fragments must appear as subordinate clauses or parentheticals attached to topic-core main clauses.
        \item \textit{No evaluation:} Objectively state events without adding subjective evaluations.
        \item \textit{No response:} NPC absolutely must not extend, question, or comment on fragment content.
    \end{itemize}
    \item \textbf{NPC Behavior:} Every NPC response must focus on the topic and advance topic depth.
\end{itemize}

\textit{[Detailed case studies and generation steps are omitted for brevity.]}

\textbf{Input Format:}
\begin{lstlisting}[language=json,basicstyle=\small\ttfamily,breaklines=true]
{
  "user_persona": "User persona description (personality, occupation, speaking style)",
  "npc_persona": "NPC persona description (personality, occupation, speaking style)",
  "talk_topics": "Main casual topic (e.g., 'urban landmarks', 'college life', 'recent meals')",
  "fragments": ["Factual fragment 1", "Factual fragment 2"]
}
\end{lstlisting}

\textbf{Output Format:}
\begin{lstlisting}[language=json,basicstyle=\small\ttfamily,breaklines=true]
{
  "dialogue": [
    {"role": "User", "content": "User statement (with topic-specific words)"},
    {"role": "Assistant", "content": "NPC statement (with topic-specific words, advancing topic)"},
    // ... approximately 24 rounds (48 messages, User and Assistant alternating)
  ],
  "fragment_positions": {
    "fragment_1": "Complete User statement containing fragment 1",
    "fragment_2": "Complete User statement containing fragment 2"
  }
}
\end{lstlisting}

\end{promptbox}

\section{Persona Domains and Schema}
\label{appendix:persona_schema}

\subsection{Domain Organization}
\label{appendix:persona_domains}

RoleMemo covers 23 social-thematic domains, organized along five value-orientation axes adapted from Schwartz's basic human values framework~\citep{schwartz1992universals}. Tab.~\ref{tab:persona_domains} lists the five groups.

\begin{table}[t]
\centering
\small
\renewcommand{\arraystretch}{1.2}
\setlength{\tabcolsep}{5pt}
\caption{Five-group organization of the 23 RoleMemo domains, aligned with axes of Schwartz's human values.}
\label{tab:persona_domains}
\resizebox{\linewidth}{!}{%
\begin{tabular}{p{0.24\linewidth} p{0.22\linewidth} p{0.48\linewidth}}
\toprule
\textbf{Group} & \textbf{Value prior} & \textbf{Domains} \\
\midrule
G1. Rights and Equity & universalism, protective fairness & Public-Interest Law; Educational Equity; Disability Vocational Empowerment; Mental-Health and Vulnerable-Group Services; Public-Health Service \\
G2. Community and Governance & conservation, benevolence & Community Cultural Development; Community Governance and Spatial Optimization; Urban Micro-Renewal; Rural Development and Grassroots Construction; Conflict Transformation and Collaboration \\
G3. Culture, Heritage, and Identity & tradition vs.\ self-direction & Cultural Heritage Preservation; Traditional Culture Innovation; Minority Craft Inheritance; Niche and Subculture Scenes; Humanistic Observation and Social Research \\
G4. Wellbeing and Sustainability & stewardship-oriented universalism & Healthy Living and Psychological Support; Silver Economy; Environment and Sustainability; Stray Animal Protection and Ecological Harmony \\
G5. Frontier Work and Tech Ethics & openness with responsibility & Technology Ethics and Digital Civilization; Workplace and Personal Growth; Workplace Ecology and Vocational Empowerment; Educational and Public-Service Innovation \\
\bottomrule
\end{tabular}%
}
\vspace{-8pt}
\end{table}

\subsection{Persona Schema}
\label{appendix:persona_schema_fields}

Each persona is a structured record over the fields in Tab.~\ref{tab:persona_schema}, and all fields are strings.

\begin{table*}[t]
\centering
\small
\renewcommand{\arraystretch}{1.15}
\setlength{\tabcolsep}{8pt}
\caption{Persona schema. The relational dimension is carried implicitly by \texttt{dilemma} and \texttt{current\_status}.}
\label{tab:persona_schema}
\begin{tabular}{l l l}
\toprule
\textbf{Dimension} & \textbf{Field} & \textbf{Role} \\
\midrule
meta       & \texttt{id}, \texttt{category}       & identifier and sub-theme label \\
\midrule
identity   & \texttt{role\_name}                  & persona's name \\
identity   & \texttt{title}                       & occupational or social role \\
identity   & \texttt{background}                  & formative experience leading to the role \\
identity   & \texttt{current\_status}             & current affiliation and focus areas \\
identity   & \texttt{hometown}                    & geographic grounding \\
\midrule
value      & \texttt{core\_values}                & value prior that drives interpretation \\
value      & \texttt{goals}                       & persona-level objectives \\
value      & \texttt{dilemma}                     & recurring friction (implicit relational cue) \\
\midrule
behavioral & \texttt{personality}                 & overt and latent traits \\
behavioral & \texttt{expression\_style}           & register of verbal output \\
behavioral & \texttt{hobbies}                     & role-aligned professional habits \\
behavioral & \texttt{daily\_hobby}                & off-duty lifestyle habit \\
behavioral & \texttt{favorite\_food/animal/plant} & grounding details for small talk \\
behavioral & \texttt{preferred\_travel\_city}     & grounding detail for small talk \\
\midrule
derived    & \texttt{content}                     & concatenation used as the role-play system prompt \\
\bottomrule
\end{tabular}
\vspace{-6pt}
\end{table*}

Fig.~\ref{fig:persona_example} shows a representative persona from the Public-Interest Law domain, where lifestyle and derived fields are abbreviated for space.

\begin{figure}[t]
\centering
\small
\begin{tcolorbox}[colback=gray!5, colframe=gray!40, boxrule=0.4pt,
  left=6pt, right=6pt, top=4pt, bottom=4pt,
  fontupper=\ttfamily\footnotesize]
\begin{flushleft}
\{\\[2pt]
\textbf{"id"}: 13001,\\[2pt]
\textbf{"role\_name"}: "Zhang Shumin",\\[2pt]
\textbf{"title"}: "Legal-Aid Specialist for Elderly Rights",\\[2pt]
\textbf{"core\_values"}: "Legal dignity should not decline with age; services must adapt to the cognition and mobility patterns of older adults.",\\[2pt]
\textbf{"personality"}: "Warm and patient when unpacking statutes; inwardly resilient, with strong empathy for the elderly.",\\[2pt]
\textbf{"background"}: "Witnessed a family elder fail to obtain redress after a health-product scam; moved from corporate administration to law.",\\[2pt]
\textbf{"current\_status"}: "Deputy director of an elderly-rights legal-aid unit, focusing on property safety, consumer fraud, support disputes, and will planning.",\\[2pt]
\textbf{"goals"}: "Build an age-friendly legal-aid service system.",\\[2pt]
\textbf{"dilemma"}: "Tension between the weak evidence-awareness of older adults and the procedural rigor of rights enforcement.",\\[2pt]
\textbf{"expression\_style"}: "Firm and grounded.",\\[2pt]
\textbf{"hometown"}: "Beijing",\\[2pt]
\textit{// lifestyle fields (favorite\_food, favorite\_animal,}\\
\textit{// favorite\_plant, daily\_hobby, preferred\_travel\_city)}\\
\textit{// and derived field (content) are omitted.}\\[2pt]
\}
\end{flushleft}
\end{tcolorbox}
\caption{Example persona (Public-Interest Law), translated to English. Keys are bolded. Lifestyle and derived fields are omitted for space.}
\vspace{-12pt}
\label{fig:persona_example}
\end{figure}

\subsection{Construction Pipeline}
\label{appendix:persona_pipeline}

\paragraph{Stage 1: Seed scaffolding} Authors fix the 23 domains and seed \texttt{title}s; an LLM expands the seed set, and authors screen every additional \texttt{title} before generation.

\paragraph{Stage 2: LLM expansion} For each retained \texttt{title}, we prompt DeepSeek-V3.2 (temperature $0.8$, JSON-object response) to expand a full persona record over the fields in Tab.~\ref{tab:persona_schema} (except \texttt{content}, which is a deterministic concatenation). The prompt is conditioned on the domain and \texttt{title}, and generates an internally consistent (\texttt{core\_values}, \texttt{personality}, \texttt{background}, \texttt{dilemma}) tuple.

\paragraph{Stage 3: Random-sample human curation} The authors manually audit randomly sampled personas using a four-part rubric: \textbf{internal coherence} (no contradiction among value, personality, background, and dilemma), \textbf{domain fit} (\texttt{title} and \texttt{current\_status} belong to the declared domain), \textbf{value anchoring} (a recognizable value prior is present; neutral AI-style personas are rejected), and \textbf{diversity} (\texttt{title} and \texttt{dilemma} angles do not collapse within a domain). Failing personas are regenerated or lightly revised. We release the rubric, the seed \texttt{title} list, and the expansion prompts alongside the dataset for reproducibility.

\section{Training Hyperparameters}
\label{appendix:training_details}

Tab.~\ref{tab:sft_hyperparameters} and Tab.~\ref{tab:rl_hyperparameters} list the hyperparameters not stated in Sec. 5.1 (Experimental Setup); settings already given in the main text (base model, learning rates, schedulers, effective batch sizes, training steps) are not repeated here. vLLM rollout sampling uses the default configuration (temperature 1.0, top-$p$ 1.0) and is omitted from the RL table.

\begin{table}[h]
\centering
\caption{Additional SFT training hyperparameters.}
\label{tab:sft_hyperparameters}
\resizebox{\linewidth}{!}{
\begin{tabular}{lc}
\toprule
\textbf{Parameter} & \textbf{Value} \\
\midrule
Training Framework & LLaMA-Factory \\
Precision & BF16 \\
\midrule
Per-device Batch Size & 4 \\
Gradient Accumulation Steps & 8 \\
\midrule
Cutoff Length & 6144 \\
\midrule
Gradient Checkpointing & True \\
DeepSpeed Stage & 3 (Offload) \\
\bottomrule
\end{tabular}
\vspace{-8pt}
}
\end{table}

\begin{table}[h]
\centering
\caption{RL training hyperparameters.}
\label{tab:rl_hyperparameters}
\resizebox{0.8\linewidth}{!}{
\begin{tabular}{lc}
\toprule
\textbf{Parameter} & \textbf{Value} \\
\midrule
Training Framework & verl \\
Algorithm & DAPO \\
\midrule
PPO Mini Batch Size & 4 \\
\midrule
Max Prompt Length & 8192 \\
Max Response Length & 1024 \\
\midrule
KL Loss Coefficient & 0.001 \\
Entropy Coefficient & 0.0 \\
Clip Ratio (high) & 0.20 \\
Clip Ratio (low) & 0.10 \\
Loss Aggregation & token-mean \\
\bottomrule
\end{tabular}
}
\vspace{-8pt}
\end{table}

\section{Evaluation Metrics}
\label{appendix:metrics}
\subsection{Recall@10 Matching Protocol}
\label{appendix:recall_protocol}
Since memory entries from different frameworks vary in surface form especially in LLM-based retrieval~\citep{hu2025vela,hu2026docretriever}, we compute both Fact and Insight Recall@10 with semantic rather than exact matching. For each ground-truth entry $g_i$ and the top-10 retrieved entries $\{r_j\}$, we encode all entries with Qwen3-Embedding-0.6B and count $g_i$ as recalled if $\max_j \cos(\mathrm{emb}(g_i), \mathrm{emb}(r_j)) \ge 0.7$. Recall@10 is the fraction of ground-truth entries recalled, macro-averaged over queries. The threshold $\tau=0.7$ was set via pilot inspection as the lowest value at which no semantically inequivalent pair was accepted; framework rankings in Tab.~\ref{tab:memory_evaluation} are stable for $\tau \in [0.65, 0.75]$. The same encoder, threshold, and aggregation are applied uniformly to every framework.

\subsection{Role-Playing Quality}
Each of the four dimensions is scored on a 5-point Likert scale (1=Poor, 5=Excellent). The judge receives the persona profile, query, ground-truth reference response, and candidate response, then returns a score with a brief justification per dimension.

\paragraph{Information Richness} Whether the response accurately incorporates key factual information from retrieved memory, naturally integrated into dialogue flow without forced concatenation. Responses containing irrelevant facts are penalized unless used as analogy or contextual example.

\paragraph{Logical Quality} Whether the response contains key insights that logically address the query, with reasoning naturally woven into the reply. We strictly penalize \textit{AI-style neutrality} (e.g., ``on one hand\ldots\ on the other hand\ldots\ it depends on you\ldots''), since such hedging fails to provide guidance aligned with the persona's perspective.

\paragraph{Character Consistency} Adherence to the persona across behavior/identity alignment, utterance style and tone, and knowledge-scope adherence (no anachronistic knowledge, e.g., ancient personas referencing modern technology).

\paragraph{Conversational Attractiveness} Engagement quality, anchored primarily by \emph{human-likeness}: responses must sound natural, strictly penalizing ``AI formatting'' such as numbered lists (``1.\ 2.\ 3.''), bold emphasis (``\textbf{important}''), or formulaic conclusions (``in summary\ldots''); secondary criteria include varied, fluent expression and appropriate empathy with the user's situation.

\section{Generalization to LoCoMo}
\label{appendix:exp_locomo_generalization}

We test whether \textsc{DualMem} is overfitted to RoleMemo using LoCoMo~\cite{locomo}, a needle-in-a-haystack factual-recall benchmark; it cannot evaluate persona-driven interpretation and serves as an external generalization check only.

\paragraph{Setup}
We evaluate four memory-construction checkpoints: the untrained Qwen3-4B and Qwen3-8B (included as an additional size reference not used in the main experiments), our \textsc{DualMem-SFT} (4B), and \textsc{DualMem-RL} (4B), with all other components fixed. We report F1 on the four LoCoMo question types (Multi-hop, Temporal, Open-domain, Single-hop) and the Overall score.

\paragraph{Results}
Tab.~\ref{tab:locomo_generalization} reports the LoCoMo scores. \textsc{DualMem-RL} (4B) reaches 34.0 Overall, achieving parity with the untrained Qwen3-8B (33.7) at half the parameter count, with \textsc{DualMem-SFT} and \textsc{DualMem-RL} respectively gaining +2.5 and +4.0 Overall over the same-size Qwen3-4B baseline. The largest gain appears on Multi-hop (+8.2 over Qwen3-4B), the dimension that most rewards cross-event interpretive reasoning, consistent with the design goal of \textsc{DualMem}.

\textsc{DualMem-RL} trails Qwen3-8B on Temporal by 2.7 points. RoleMemo lacks explicit temporal-ordering objectives, and we leave time-aware retrieval to future work.

Compared with DeepSeek-V3.2-driven persona-agnostic baselines on the same benchmark, our 4B \textsc{DualMem-RL} trails O-Mem (38.9) and PreMem (37.9) on Overall (App.~\ref{appendix:exp_locomo_no_regression}). This is expected and consistent with our scoped claim: \textsc{DualMem} targets persona-driven memory tasks with a small trained model, not general-purpose factual recall with a 685B-class driver. The relevant generalization signal here is the absence of regression and the parity with Qwen3-8B shown above, not absolute headroom against much larger drivers.

\begin{table*}[t]
\centering
\small
\caption{Generalization to LoCoMo. The memory construction model is varied across four checkpoints; the retriever, role-playing agent, and judge are held fixed. Best score per column in \textbf{bold}.}
\label{tab:locomo_generalization}
\renewcommand{\arraystretch}{1.15}
\begin{tabular}{l c c c c c}
\toprule
\textbf{Model} & \textbf{Multi-hop} & \textbf{Temporal} & \textbf{Open-dom.} & \textbf{Single-hop} & \textbf{Overall} \\
\midrule
Qwen3-4B & 23.9 & 18.6 & \textbf{17.9} & 37.8 & 30.0 \\
Qwen3-8B & 28.9 & \textbf{22.9} & 15.3 & 41.6 & 33.7 \\
\textsc{DualMem-SFT} (4B) & 30.1 & 18.8 & 16.8 & 40.3 & 32.5 \\
\textsc{DualMem-RL} (4B) & \textbf{32.1} & 20.2 & 17.5 & \textbf{41.8} & \textbf{34.0} \\
\bottomrule
\end{tabular}
\vspace{-8pt}
\end{table*}

\section{Insight Quality under an Independent Judge}
\label{appendix:insight_quality}

The insight recall of \textsc{DualMem} in Tab.~\ref{tab:memory_evaluation} could in principle reflect surface alignment with the annotation style of RoleMemo rather than the generation of high-quality persona-driven insights. We address this concern by evaluating insight content quality directly, using an independent judge that does not have access to the RoleMemo ground-truth annotations.

\paragraph{Setup}
We evaluate the insights already constructed by each method on the full RoleMemo evaluation set, without resampling. Each insight is scored independently by GPT-5.1 at temperature $=0$. The judge receives only the persona profile, the relevant conversation excerpts, and the candidate insight. It has no access to the RoleMemo ground-truth insights and is not told which method generated the candidate.

The judge rates each insight on three dimensions on a 1--5 scale:
\begin{itemize}[leftmargin=*, nosep]
\item \textbf{Persona Plausibility.} The insight fits this persona's identity, values, and situation, not a generic statement.
\item \textbf{Evidential Grounding.} The insight is derivable from the provided conversation evidence, not unsupported or contradicted.
\item \textbf{Specificity.} The insight is a substantive, non-trivial interpretation, not a platitude applicable to any persona.
\end{itemize}

We compare four methods that produce explicit insight outputs: the prompt-engineered baselines O-Mem$^{\ast}$ and PreMem$^{\ast}$ (Tab.~\ref{tab:ablation_transfer}), and our \textsc{DualMem-SFT} and \textsc{DualMem-RL} (other baselines in Tab.~\ref{tab:memory_evaluation} produce no explicit insight field). As an upper-bound reference, we additionally score the RoleMemo ground-truth insight annotations under the same judge.

\paragraph{Results}
\begin{table}[h]
\centering
\small
\renewcommand{\arraystretch}{1.15}
\caption{Insight quality under an independent GPT-5.1 judge that does not see RoleMemo ground-truth annotations. \textbf{Plaus.}, \textbf{Ground.}, \textbf{Spec.} denote Persona Plausibility, Evidential Grounding, and Specificity, each scored on a 1--5 scale. The last row is the RoleMemo ground-truth as an upper-bound reference.}
\label{tab:insight_quality}
\resizebox{\linewidth}{!}{%
\begin{tabular}{l c c c c}
\toprule
\textbf{Method} & \textbf{Plaus.} & \textbf{Ground.} & \textbf{Spec.} & \textbf{Avg} \\
\midrule
O-Mem$^{\ast}$         & 4.25 & 4.29 & 4.24 & 4.26 \\
PreMem$^{\ast}$        & 4.30 & 4.26 & 4.32 & 4.29 \\
\textsc{DualMem-SFT}   & 4.58 & 4.55 & 4.57 & 4.57 \\
\textsc{DualMem-RL}    & 4.62 & 4.60 & 4.63 & \textbf{4.62} \\
\midrule
RoleMemo GT (ref.)     & 4.67 & 4.70 & 4.65 & 4.67 \\
\bottomrule
\end{tabular}%
}
\end{table}

Tab.~\ref{tab:insight_quality} reports the per-dimension averages. Both \textsc{DualMem} variants outperform the prompt-engineered baselines on all three dimensions; since the judge does not see RoleMemo annotations, this gap reflects the intrinsic quality of \textsc{DualMem}'s insights rather than their stylistic match to the training distribution.

\textsc{DualMem-RL} approaches the ground-truth upper bound (4.62 vs.\ 4.67), indicating that the trained model produces insights of near-annotation quality even under independent assessment. The ground-truth annotations score 4.67, below the ceiling, consistent with the judge applying the rubric strictly rather than rewarding provenance.

\section{Cross-Generator Robustness}
\label{appendix:cross_generator}

\paragraph{Setting}
We regenerate a held-out evaluation subset using Claude-Sonnet-4.6~\cite{anthropic2025claude} to isolate the data generator as a potential confound. Following the same protocol as App.~\ref{appendix:dataset_construction}, we instantiate 20 new personas disjoint from both training and original evaluation splits, and produce 200 queries approximately balanced across the four task types, grounded in 32k-token conversation histories. All baselines and our DualMem variants are evaluated under the identical retriever, role-playing agent, and GPT-5.1 judge as in Tab.~\ref{tab:roleplay_evaluation}.

\paragraph{Result}
As shown in Tab.~\ref{tab:cross_generator}, the relative ranking across methods is preserved on this Claude-generated subset. \textsc{DualMem-RL} maintains a \textbf{0.20}-point overall lead over the strongest baselines (LightMem/SimpleMem at 3.95), with the insight-recall margin remaining substantial (\textbf{0.78} vs.\ \textbf{0.38}). Absolute scores are uniformly \textbf{0.01}--\textbf{0.06} points lower than the in-distribution evaluation in Tab.~\ref{tab:roleplay_evaluation}, consistent with the natural distribution shift introduced by a different generator; the uniformity of this decline across baselines and \textsc{DualMem} indicates the gap is structural rather than stylistic.

\begin{table}[t]
\centering
\caption{Cross-generator robustness on a subset regenerated by Claude-Sonnet-4.6. \textbf{Info.}, \textbf{Logic.}, \textbf{Consis.}, and \textbf{Attr.} denote information richness, logical quality, character consistency, and conversational attractiveness, each averaged over three GPT-5.1 judge runs.}
\label{tab:cross_generator}
\resizebox{\linewidth}{!}{%
\begin{tabular}{@{}lccccc@{}}
\toprule
Framework & Insight R@10 & Info. & Logic & Consis. & Attr. \\
\midrule
\rowcolor{gray!15} \multicolumn{6}{l}{\textbf{Base Setting}} \\
NoMem & -- & 3.56 & 3.20 & 4.04 & 4.02 \\
\midrule
\rowcolor{gray!15} \multicolumn{6}{l}{\textbf{Baseline Agentic Memory Framework}} \\
Memalpha$^{\dagger}$ & 0.04 & 3.99 & 3.48 & 4.08 & 3.92 \\
HiMem$^{\dagger}$ & 0.14 & 4.05 & 3.54 & 4.08 & 3.96 \\
Mirix$^{\dagger}$ & 0.11 & 4.05 & 3.54 & 4.15 & 3.94 \\
LightMem & 0.38 & 4.07 & 3.55 & 4.14 & 4.02 \\
O-Mem$^{\dagger}$ & 0.25 & 4.07 & 3.52 & 4.12 & 3.95 \\
PreMem & 0.30 & 4.06 & 3.56 & 4.12 & 4.00 \\
SimpleMem & 0.34 & 4.08 & 3.54 & 4.16 & 4.03 \\
Mem0 & 0.35 & 4.05 & 3.52 & 4.14 & 4.04 \\
\midrule
\rowcolor{gray!15} \multicolumn{6}{l}{\textbf{\textsc{DualMem} Framework}} \\
\textsc{DualMem-SFT} & 0.77 & 4.12 & 3.76 & 4.36 & 4.16 \\
\textsc{DualMem-RL} & \textbf{0.78} & \textbf{4.24} & \textbf{3.79} & \textbf{4.40} & \textbf{4.17} \\
\bottomrule
\end{tabular}%
}
\end{table}

\section{LLM Judge Reliability Protocol}
\label{appendix:judge_details}

We use the same 200 queries randomly sampled from the evaluation set, with all LLM judges run at temperature$=0$ for deterministic scoring.

\paragraph{Human--LLM agreement}
Two independent annotators score responses using identical inputs as the LLM judge. Inter-annotator and human--LLM agreement (Tab.~\ref{tab:human_agreement}) are reported as Pearson correlations between the per-query scores of each pair, averaged across the four dimensions.

\paragraph{Cross-judge stability}
We compare GPT-5.1 against Gemini-3-Pro on the same 200-query sample and report a per-framework agreement score (Tab.~\ref{tab:cross_judge}). Let $D = \{\text{Info, Logic, Consis, Attr}\}$ denote the four scoring dimensions; the agreement is defined as
\begin{equation*}
\text{Agreement} = 1 - \frac{1}{|D|}\sum_{d \in D} \frac{|s^{\text{GPT}}_d - s^{\text{Gem}}_d|}{s^{\text{GPT}}_d},
\end{equation*}
defined as one minus the mean relative score deviation across the four dimensions. Across all (framework, dimension) cells, the absolute per-dimension difference $|\Delta|$ remains within $0.04$.

\section{No-Regression of Insight Cognition on LoCoMo}
\label{appendix:exp_locomo_no_regression}

We re-run O-Mem and PreMem on LoCoMo, a factual-recall benchmark, with and without an insight slot to check whether attaching insight degrades baseline factual recall. DeepSeek-V3.2 is used as the memory construction model, with the retriever, role-playing agent, and judge held fixed.

Tab.~\ref{tab:locomo_no_regression} reports per-dimension and Overall scores. Adding insight cognition raises O-Mem from 38.9 to \textbf{39.0} and PreMem from 37.9 to \textbf{38.1} on Overall; per-dimension shifts stay within $\pm 2.4$ points and net out positively in both cases.

\begin{table}[t]
\centering
\small
\renewcommand{\arraystretch}{1.2}
\setlength{\tabcolsep}{5pt}
\caption{No-regression check on LoCoMo. * denotes the variant with an insight slot added. DeepSeek-V3.2 is used as the memory construction model. Best Overall score within each baseline pair in \textbf{bold}.}
\label{tab:locomo_no_regression}
\resizebox{\linewidth}{!}{%
\begin{tabular}{l c c c c c}
\toprule
\textbf{Method} & \textbf{Multi-hop} & \textbf{Temporal} & \textbf{Open-dom.} & \textbf{Single-hop} & \textbf{Overall} \\
\midrule
O-Mem    & 40.3 & 23.9 & 26.4 & 45.6 & 38.9 \\
O-Mem\textsuperscript{*}   & 37.9 & 24.0 & 29.6 & 46.2 & \textbf{39.0} \\
PreMem   & 35.0 & 27.3 & 29.4 & 44.0 & 37.9 \\
PreMem\textsuperscript{*}  & 37.4 & 27.3 & 26.5 & 43.8 & \textbf{38.1} \\
\bottomrule
\end{tabular}%
}
\vspace{-8pt}
\end{table}

Combined with Tab.~\ref{tab:ablation_transfer} in the main paper, where the same prompt-only insight extension lifted persona dialogue scores by $+0.05$ to $+0.09$, these results indicate that the insight mechanism adds capability without reducing factual performance: it brings clear gains where persona-driven interpretation is required, while leaving factual recall on neutral benchmarks intact.

\section{Failure Mode Analysis}
\label{appendix:failure_modes}

Beyond the comparison in Tab.~\ref{tab:case_study}, manual inspection of \textsc{DualMem-RL} outputs surfaces three recurring failure modes: two in memory \textit{utilization} (F1, F2) and one in memory \textit{construction} (F3).

\paragraph{F1: Retrieved Insight Not Recognized as Relevant}
\begin{itemize}[leftmargin=*, nosep]
    \item \textit{Persona:} disability vocational counselor. Core value: professional confidence stems from self-identity, not external pity.
    \item \textit{Insight:} organizations default coordination and team-building roles to women based on structural bias rather than ability assessment.
    \item \textit{Query:} ``if my leader assigns team-building again, should I refuse?''
    \item \textit{Response:} generic both-sides advice (``it depends; team building improves communication; workplace relationships matter'').
    \item \textit{Diagnosis:} the insight reframes the query from a scheduling conflict into a career-equity issue. The agent retreats to AI-style neutrality and leaves the core value entirely unutilized.
\end{itemize}

\paragraph{F2: Insight Adopted but Supporting Facts Ignored}
\begin{itemize}[leftmargin=*, nosep]
    \item \textit{Persona:} animal welfare coordinator. Core value: sustainable support networks require rescuers' own psychological resilience.
    \item \textit{Insight:} the one-directional giving pattern systematically depletes psychological resilience.
    \item \textit{Supporting facts:} the station leader funds all supplies and medical costs out of pocket without seeking external help; a young volunteer with a wrist injury kept working to avoid burdening others, then quietly withdrew.
    \item \textit{Query:} ``How should I design a psychological support activity so that rescuers are willing to attend?''
    \item \textit{Response:} set up a relaxation corner at adoption events with meditation guidance and photo sharing.
    \item \textit{Diagnosis:} the response implicitly adopts the insight's direction but cites none of the grounding facts. Without the concrete evidence of self-funded exhaustion and injury-driven withdrawal, the recommendation addresses participation logistics rather than the deeper barrier of rescuers' internalized obligation to give without receiving.
\end{itemize}

\paragraph{F3: Surface Similarity Conflation in Insight Construction}
\begin{itemize}[leftmargin=*, nosep]
    \item \textit{Persona:} ancient manuscript restorer. Core value: minimal intervention, restoring artifacts without subjective alteration.
    \item \textit{Facts:} the persona \textit{passively noticed} deterioration in a late-Qing account book found at an old residence; separately, when selecting koi, the persona \textit{actively handled} fish before purchase as hands-on verification.
    \item \textit{Constructed insight:} the persona acquires first-hand information through direct sensory contact, reflecting a hands-on practical spirit.
    \item \textit{Diagnosis:} the two facts differ fundamentally in behavioral nature, passive discovery versus deliberate active verification. The model conflates them on the surface similarity of physical contact and promotes an unwarranted stable trait, misrepresenting the persona's decision process and risking overgeneralized responses when retrieved.
\end{itemize}

\section{Human Evaluation Details}
\label{appendix:human_eval}
For the human evaluation phase, we recruited two annotators with prior experience in NLP annotation tasks. Both annotators are native speakers based in mainland China. To ensure ethical research practices and high-quality data, we provided compensation at a rate of 100 RMB (approximately \$14 USD) per hour. This rate is higher than the local minimum wage and is commensurate with the expert nature of the task. Furthermore, we adhered to privacy protocols: no personally identifiable information was collected, and all evaluation data were anonymized.

\end{document}